\documentclass[11pt, a4paper, onecolumn]{article}
\usepackage[margin=1in]{geometry}
\usepackage[utf8]{inputenc}
\usepackage[T1]{fontenc}
\usepackage{lmodern}
\usepackage{microtype}
\usepackage{amsmath, amssymb, amsfonts}
\usepackage{graphicx}
\usepackage{url}
\usepackage[colorlinks=true, allcolors=blue]{hyperref}
\usepackage{authblk}
\usepackage{multicol}
\usepackage[authoryear, sort&compress, round]{natbib}
\title{Measuring Progress Toward AGI: A Cognitive Framework}

\author[1]{Ryan Burnell\thanks{Corresponding author: rburnell@google.com}}
\author[1]{Yumeya Yamamori}
\author[1]{Orhan Firat}
\author[1]{Kate Olszewska}
\author[1]{Steph Hughes-Fitt}
\author[1]{Oran Kelly}
\author[1]{Isaac R. Galatzer-Levy}
\author[1]{Meredith Ringel Morris}
\author[1]{Allan Dafoe}
\author[1]{Alison M. Snyder}
\author[1*]{Noah D. Goodman}
\author[1*]{Matthew Botvinick}
\author[1]{Shane Legg}

\affil[1]{Google DeepMind}
\affil[*]{Work done while at Google DeepMind.}

\begin{document}

\maketitle

\begin{abstract}
Despite widespread discussion of AGI, there is no clear framework for measuring progress toward it. This ambiguity fuels subjective claims, makes it difficult to track progress, and risks hindering responsible governance. As a starting point to address this gap, we present a framework for understanding system capabilities in relation to human cognitive abilities. Drawing from decades of research in psychology, neuroscience, and cognitive science, we introduce a Cognitive Taxonomy that deconstructs general intelligence into 10 key cognitive faculties. We then propose a rigorous evaluation protocol in which a system's performance is measured across a suite of targeted, held-out cognitive tasks, generating a `cognitive profile' that can be used to understand a system's strengths and weaknesses. We hope this framework will provide a practical roadmap and an initial step toward more rigorous, empirical evaluation of AGI.
\end{abstract}

\section{Introduction}
Artificial general intelligence (AGI) has the potential to accelerate scientific discovery, increase productivity and help solve some of humanity's most pressing problems. Yet our ability to understand how close we are to this critical milestone is hampered by a lack of clarity about how general intelligence should be operationalized and measured. As a result, the capabilities of today's AI tools are often underestimated or overhyped while the potential benefits and risks of future systems remain poorly understood. This lack of empirical grounding makes it difficult for researchers to communicate progress and for policymakers to craft effective governance, ultimately hindering our collective ability to navigate the path to AGI responsibly. To address this critical gap, we introduce a cognitive framework for measuring progress on increasingly capable AI systems.

\subsection{Operationalizing AGI}
The term AGI has a long history. It was first used by Mark Gubrud in a 1997 paper to refer to systems that \textit{"rival or surpass the human brain in complexity and speed, that can acquire, manipulate and reason with general knowledge, and that are usable in essentially any phase of operations where a human intelligence would otherwise be needed"} \citep{gubrud1997nanotechnology}. Then, in 2001, without knowledge of Gubrud's work, Shane Legg independently coined the term "Artificial General Intelligence" for Ben Goertzel's subsequent book, which adopted it as its title \citep{goertzel2007artificial}. There, AGI was referred to as \textit{"AI systems that possess a reasonable degree of self-understanding and autonomous self-control, and have the ability to solve a variety of complex problems in a variety of contexts, and to learn to solve new problems that they didn't know about at the time of their creation"}. Since then, AGI has become a mainstay in discussions about AI capabilities. However, the term is often used as a shorthand to describe various kinds of highly capable AI systems. Given the important societal and scientific implications of generally capable AI systems, it is vital that we establish precise and robust ways to measure progress toward this milestone.

As a first step in this direction, in 2023 Google DeepMind presented the \textit{Levels of AGI} framework that proposed a series of important stages on the path toward general intelligence \citep{morris2024position}. The framework considers intelligence a continuous, multidimensional construct and argues it is important for systems to be both highly \textit{capable} and highly \textit{general}. On both these dimensions, it is clear that human capabilities are a key reference point. The creation of a system that is capable of exhibiting all the cognitive capabilities that humans have would mark a historic moment and a philosophical milestone. Moreover, such a system would open up countless applications, including universal assistants, personalized learning, and powerful new scientific tools. 

What remains less clear is how we can understand how far away AI systems are from matching these cognitive capabilities. Over the past few years, several benchmarks aimed at measuring progress toward AGI have been proposed (see, for example, \citealt{chollet2019measure} and \citealt{hendrycks2025definition}). However, existing efforts fail to cover the full breadth of human cognition and lack robust comparisons to human performance. Here, we aim to address this gap in two parts. First, we propose a Cognitive Taxonomy that captures the important aspects of human cognition that an AGI system should be able to match. Second, we describe a framework for evaluating AI systems across this cognitive space to help us better contextualize system capabilities.

\section{Cognitive Taxonomy}
\label{sec:taxonomy-overview}
To understand where AI systems stand relative to human cognitive capabilities, we first need to identify the key cognitive processes that enable people to navigate the complex and changing world. In this endeavor, we can turn to the decades of research into human cognition---in particular, research from \textit{psychology, neuroscience, and cognitive science}, which have built and refined theories of cognition over many decades through iterative experimentation. These disciplines employ a wide range of methods, including experimental paradigms, brain imaging techniques, patient studies, and computational modeling, providing rich and empirically grounded insights we can draw from. Here, we use these insights to create a \textit{cognitive taxonomy} describing the cognitive capabilities that evidence suggests are important for general intelligence.
 
Of course, characterizing intelligence is far from a simple task. Even within the field of human cognitive science there are many debates that remain unsettled and many questions that remain unanswered \citep{stainton2006contemporary}. The world of artificial systems is perhaps even more complex, rich with a multitude of architectures and training algorithms improving at a pace that makes evolution pale in comparison. It is therefore possible---even likely---that some aspects of human cognition may not be relevant in the context of artificial systems. Conversely, we are likely to find cognitive processes in these artificial systems that humans do not possess.  

For this reason, our goal was not to create \textit{the} definitive account of all of cognition, but rather to build a practical framework for evaluation that is both theoretically grounded and comprehensive enough to cover the breadth of human intelligence. We consider this framework a starting point, and we hope it can provide a foundation for a robust science of artificial general intelligence.

\subsection{Cognitive faculties}
\begin{figure}[htbp]
    \centering
    \includegraphics[width=0.8\columnwidth]{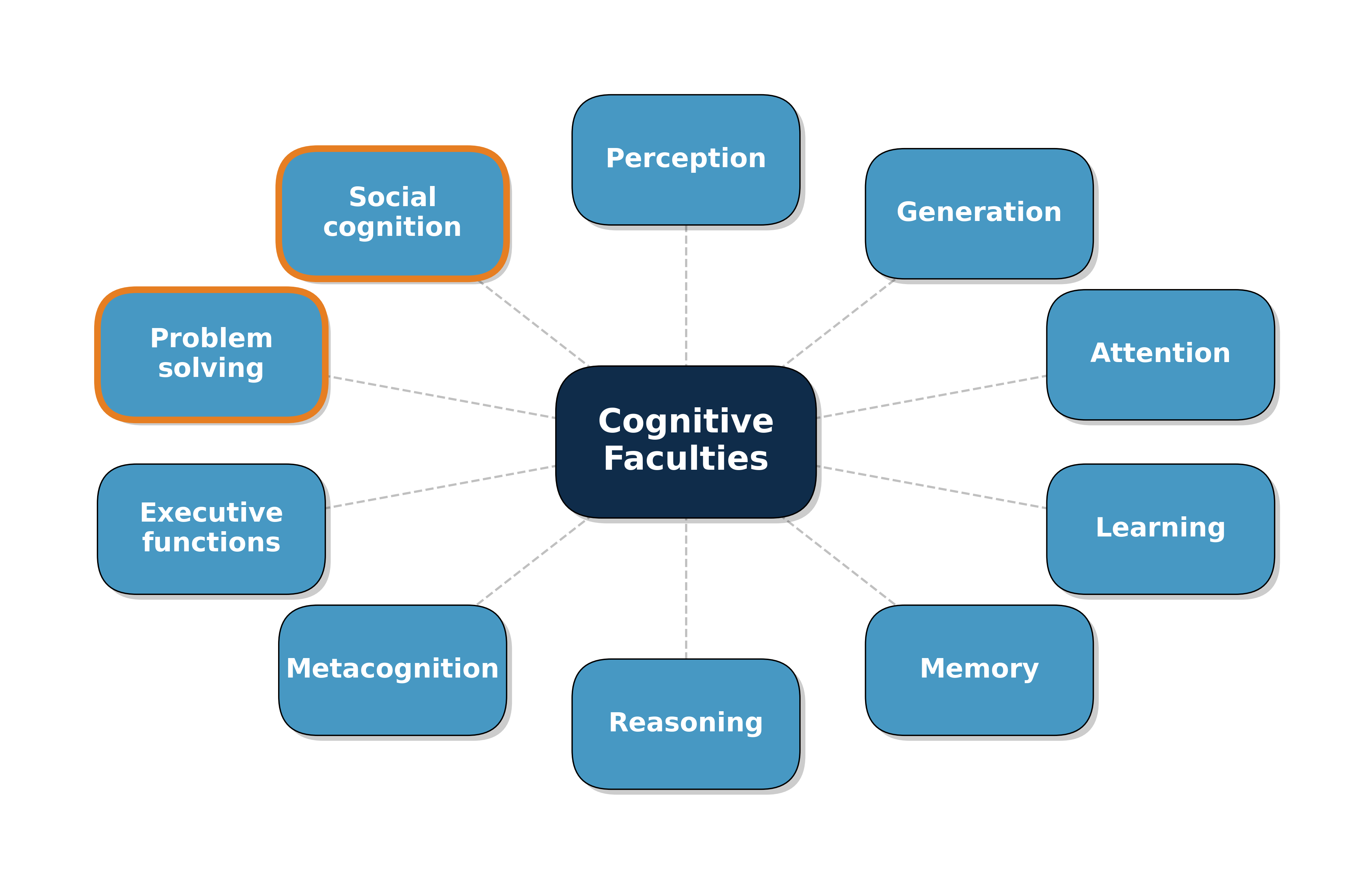}
    \caption{\small\it Overview of the 10 cognitive faculties. Faculties outlined in orange represent composite faculties.}
    \label{fig:cognitive_taxonomy}
\end{figure}

Our cognitive taxonomy enumerates 10 cognitive faculties that the scientific efforts in mapping human cognition suggest are important for intelligent behavior (see \autoref{fig:cognitive_taxonomy}).  For each faculty and process, we identify a set of specific abilities and sub-abilities that a generally intelligent system should be expected to exhibit. For brevity, we provide brief descriptions of each faculty below and include the full taxonomy in the \hyperref[sec:full-taxonomy]{Appendix}. 

An important characteristic of our taxonomy is that it focuses on \textit{what} the system is able to accomplish not on \textit{how} it does so (see \citealt{marr1982vision} for a discussion of this distinction). In doing so, we are able to remain agnostic to the underlying mechanisms employed by a system and do not prescribe any specific modeling approaches.

We begin with eight faculties capturing the basic building blocks of human cognition:

\begin{description}
    \item[\textbf{Perception:}] The ability to extract and process sensory information from the environment.
    \item[\textbf{Generation:}] The ability to produce outputs such as speech, text, motor movements, and computer control actions.
    \item[\textbf{Attention:}] The ability to focus cognitive resources on specific aspects of perceptual stimuli, thoughts, or task demands.
    \item[\textbf{Learning:}] The ability to acquire new knowledge, skills, or understanding through experience, study, or instruction.
    \item[\textbf{Memory:}] The ability to store and retrieve information over time.
    \item[\textbf{Reasoning:}] The ability to draw valid conclusions and make inferences by applying logical principles.
    \item[\textbf{Metacognition:}] The knowledge a system has about its own cognitive processes and its ability to monitor and control those processes.
    \item[\textbf{Executive functions:}] Abilities that facilitate goal-directed behavior. Includes planning, inhibition, and cognitive flexibility.
\end{description}

These eight faculties form the basic building blocks, but they do not operate in isolation---on the contrary, they interact, overlap, work together, and build on one another \citep{kovacs2016process}. To fully understand the capabilities of a system, we also need to understand how well it can combine and apply multiple faculties in unison. We therefore propose an additional two composite faculties capturing two critical psychological contexts in which the various faculties are applied together:

\begin{description}
    \item[\textbf{Problem solving:}] The ability to find effective solutions to domain-specific problems.
    \item[\textbf{Social cognition:}] The ability to process and interpret social information and to respond appropriately in social situations.
\end{description}

We hypothesize that these ten faculties capture the key capabilities needed for a system to be both highly general and highly capable. A system with significant weaknesses in one or more of these ten faculties is likely to be unable to perform some real-world tasks that most humans can perform.

\section{Evaluating Cognitive Capabilities}
To understand the cognitive capabilities of an AI system, we need to robustly evaluate the system across each cognitive faculty and compare the system to a meaningful human baseline. We therefore propose the following three-stage evaluation protocol:

\begin{enumerate}
    \item \textbf{Conduct cognitive assessment} of the system across a broad suite of cognitive tasks.
    \item \textbf{Collect human baselines} on the same cognitive tasks to establish a point of comparison.
    \item \textbf{Build cognitive profiles} to map the system's strengths and weaknesses in relation to human performance.
\end{enumerate}

\subsection{Conduct cognitive assessment}
\label{sec:cognitive-assessment}
The first step in evaluating progress toward AGI is to \textbf{evaluate system performance on a broad suite of cognitive tasks} covering each faculty. The cognitive tasks should be:
    \begin{itemize}
        \item \textbf{Targeted to specific cognitive abilities}. It is important to include tasks that isolate each cognitive faculty in order to precisely diagnose model weaknesses. 
        \item \textbf{Held out}. Evaluations should ideally use private, held-out test sets to prevent contamination---if systems had previously seen solutions or strategies for the specific test items, then performance on those tests is unlikely to be indicative of general intelligence (e.g., see \citealt{jacovi2023stop}). 
        \item \textbf{Independently verified}. To ensure the community can be confident in the findings, both the cognitive tasks and the evaluation results should be audited by an independent third party.
        \item \textbf{Varied in difficulty for humans}. Some tasks are easy for humans and hard for AI systems (see \citealt{chollet2025arc}), so it is important to include such tasks as well as tasks that test the limits of human capabilities.
        \item \textbf{Varied in structure and format}. Idiosyncrasies in the structure of specific tasks can artificially inflate (or inhibit) performance on those tasks (this is true both for humans and AI systems; e.g. see \citealt{Cheng1985391, dasgupta2024languagemodelshumanlikecontent}). For this reason, the evaluation of each faculty should include multiple tasks with a variety of structures and formats (e.g. multiple choice vs open response, text inputs vs multimodal, multi-step vs single-turn).
    \end{itemize}

\subsection{Collect human baselines}
\label{sec:human-baselines}
To understand how close a system's capabilities are to human-level on a set of tasks, we need to \textbf{quantify human performance} on those same tasks.

These \textit{human baselines} can be constructed by asking a large sample of humans to complete the same tasks as the AI systems. The tasks should be performed under the same conditions, including the same task instructions (and few-shot examples, if any), response format, and access to external tools.

Because we want to understand the full range of human capabilities, we think it is critical to sample widely from the human population. At the same time, we want to understand how well systems will perform in real-world situations---situations that involve knowledge and abilities that are typically only fully developed in adulthood and typically honed through formal education. Therefore, we propose that a reasonable human baseline should consist of a \textbf{demographically representative} sample of \textbf{adults} with at least the equivalent of an \textbf{upper secondary education}\footnote{Equivalent to earning a high-school degree in the US.}. 

\subsection{Build cognitive profiles}
Using the evaluation results from the \hyperref[sec:cognitive-assessment]{cognitive assessment} and the \hyperref[sec:human-baselines]{human baselines}, we can next \textbf{build cognitive profiles} for each system to map out the strengths and weaknesses of the system relative to human performance across the 10 cognitive faculties.

\begin{figure}[htbp]
    \centering
    \includegraphics[width=0.9\columnwidth]{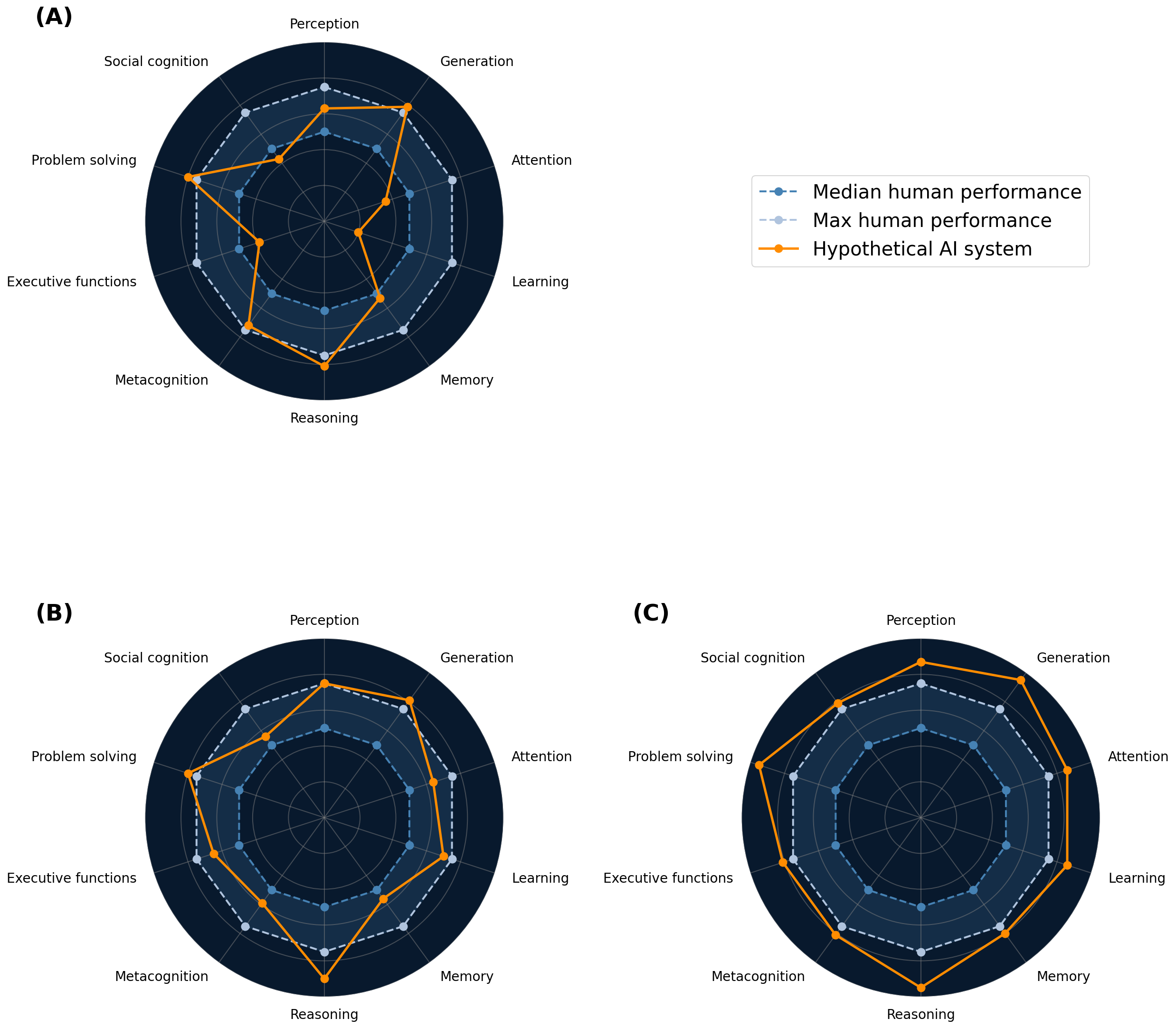}
    \caption{\small\it Cognitive profiles for three hypothetical systems. Panel A: A hypothetical system that shows significant cognitive weaknesses relative to the human sample. Panel B: A hypothetical system that outperforms the human sample median across all cognitive faculties. Panel C: A hypothetical system that outperforms the human sample maximum across all cognitive faculties. The hypothetical human sample scores are standardized for each dimension for illustrative purposes.}
    \label{fig:cognitive_profiles}
\end{figure}

To do so, we can calculate the percentage of people from the human sample the system outperforms, and place the system along the distribution of human performance (see \autoref{fig:cognitive_profiles} for examples).

Given the jagged nature of system capabilities \citep{jaggednessPreprint}, many kinds of profiles are possible. For example:
\begin{enumerate}
    \item A system scores \textbf{below the human baseline sample median} for \textbf{one or more} cognitive faculties. Such a system shows significant cognitive weaknesses and is likely to struggle in at least some real-world contexts.
    \item  A system scores \textbf{above the human baseline sample median} across \textbf{all ten} cognitive faculties. Such a system is demonstrating the ability to match the cognitive capabilities of at least 50\% of the human sample, and is likely to perform well in many real-world contexts. 
    \item A system scores in the \textbf{99th percentile} across \textbf{all ten} cognitive faculties (\hyperref[fig:cognitive_profiles]{Figure 2C}). Such a system is demonstrating it can match almost anyone from the human sample across the entire cognitive space.\footnote{A somewhat stronger test would be to require the system to score at or above the sample maximum across every faculty. But it is difficult to quantify the uncertainty around a maximum value.} Of course, any practical human baseline sample---even a highly representative one---is unlikely to capture the full scope of human capabilities, so further scrutiny would be required before drawing any conclusions based on this pattern.
\end{enumerate}

Some important methodological notes are warranted here: To construct a cognitive profile, we need to calculate an overall score for each person and each AI system on each faculty (see \autoref{fig:cognitive_profiles}). There are many ways these scores could be calculated. The simplest approach would be to aggregate together the metrics from each cognitive task within a faculty. But taking a more sophisticated statistical approach---such as building an Item-Response Theory model of system capabilities (see e.g., \citealt{MARTINEZPLUMED201918})---could provide more robust and informative results.

Regardless of which approach is used, it is important to quantify the uncertainty around each capability estimate. Uncertainty in this context stems from at least three main sources:
\begin{enumerate}
    \item \textbf{Task quality}: Factors such as prompt quality and task diversity can dramatically affect how informative the results are.
    \item \textbf{Construct validity}: Cognitive evaluations are intended to provide insights into broad cognitive faculties, such as reasoning or attention. However, if the datasets do not isolate the target faculty effectively or the data are contaminated, the results may not accurately reflect a system's general capabilities.
    \item \textbf{Stochasticity}: The stochasticity of generative AI systems adds noise to evaluation results---asking a system to complete the same task multiple times can produce wildly different results across repetitions (this is particularly true of complex tasks with multiple steps).
\end{enumerate}

Characterizing this uncertainty is crucial for understanding the extent to which differences between systems or between a system and the human sample are meaningful.

\newpage
\section{Discussion}

\subsection{Building cognitive evaluations}
Here we provide a framework for measuring AI systems across the range of human cognitive capabilities. To execute on this framework, we need robust cognitive benchmarks. Many useful benchmarks already exist for some parts of the cognitive space, including problem solving \citep{phan2025humanity}, perception \citep{patraucean2023perception}, and world knowledge \citep{cheng2025factsleaderboardcomprehensivebenchmark, haas2025simpleqa}. But there are large coverage gaps in areas such as metacognition, attention, learning, and social cognition. Moreover, many of the high-quality benchmarks that do exist are fully public, so they are susceptible to data contamination and may not provide generalizable signal. For this reason, we do not think existing benchmarks are sufficient to reliably evaluate AI cognitive capabilities.

To address these gaps, we are working with the academic community to build robust, held-out evaluations that will better enable us to evaluate increasingly capable AI systems.

\subsection{Beyond cognitive faculties}
Our taxonomy focuses on cognitive capabilities. But to build a full understanding of AI capabilities and AI behavior we will need to consider---and evaluate---many other system characteristics. Below we discuss several that we think will be important, all of which can be understood in relation to a human baseline.

\subsubsection{Processing and response speed}
For a response to be adaptive or useful, correctness is not always sufficient. Often, the response must be timely, too. Take, for example, a self-driving car system. The system's reliability depends not only on its ability to identify potential hazards, but its ability to do so quickly. Even when speed is less critical, it is still important for understanding real-world utility---a system that can fix a coding bug or book a flight in one minute is likely to be much more useful than one that takes six hours to complete the task.

The construct of processing speed is at least partly cognitive \citep{danthiir2005mental}---for example, problem solving speed depends on the thinking strategies a system employs and on how well it draws connections to relevant knowledge. But speed also depends on other factors such as hardware and network speed, so it does not fit perfectly as a cognitive faculty. Regardless, response speed is a key performance metric that should be measured across the cognitive space. 

\subsubsection{System propensities}
Another important determinant of how a system will behave when deployed is its propensities (i.e. not just what the system \textit{can} do but what it will \textit{tend} to do). How willing is the system to take risks? How aligned is it with human values? What are its typical problem-solving strategies? How does it communicate and interact with people? These kinds of behavioral factors will significantly affect how safe and reliable that system is, so we need robust tools to evaluate them. A full account of system propensities is beyond the scope of this paper, but will be a critical area of study to inform deployment decisions and enable effective governance (see e.g. \citealt{romeroalvarado2026capabilitiesaintneedmeasuring, taubenfeld2026evaluatingalignmentbehavioraldispositions}).

\subsubsection{Creativity}
The human capacity for creativity has long been a topic of interest to both philosophy and cognitive science. Yet disagreements continue about how creativity should be conceptualized and measured (see e.g. \citealt{Kaufmann01072003, Boden1994What}). One common way of thinking about creative outputs is that they need to be both \textit{novel} and \textit{high-quality} \citep{Sternberg_Lubart_1998}. But whether an output is high quality is highly subjective and domain specific, especially when it has entirely novel features. Moreover, it can be argued that these characteristics are simply general features of intelligent behavior. For this reason, it may be difficult to isolate and evaluate creativity objectively in AI systems. However, we can still evaluate the cognitive processes involved in creativity. For example, creativity is often associated with cognitive flexibility (the ability to switch modes of thinking and the ability to generate a wide variety of distinct ideas), which is covered in the taxonomy as part of executive functions. In addition, the taxonomy captures world knowledge (as part of semantic memory) and problem solving, both of which are relevant to creativity \citep{Sternberg_Lubart_1998}.

\subsubsection{End-to-end deployment evaluations}
Lastly, it is important to make clear that cognitive benchmarking is not a substitute for applied, end-to-end evaluations---if a system is being deployed in a specific context, it is absolutely critical to evaluate the system on important deployment workflows. These two approaches are complementary---cognitive evaluations can help explain model failures and inform model improvements; real-world, deployment evaluations can inform deployment decisions and predict economic impacts (see, e.g. \citealt{patwardhan2025gdpvalevaluatingaimodel, mazeika2025remotelaborindexmeasuring}). 

\subsection{Model vs system evaluation}
Historically, evaluations were largely focused on a specific model checkpoint. But modern AI systems are more than just a model---they are deployed with specific system instructions, have access to tools, can manipulate their environments via actions, and may even have the ability to make calls to other AI systems. 

How should we approach evaluations in light of this added complexity? We think that attempting to isolate the core model of a system from its other components will become increasingly impractical since the inner workings of new systems are often not disclosed and these different components may not even be easily separable. This model-only approach will also become less and less informative, since the results would not be representative of how the system will operate or perform when deployed with access to these components. Modularity is also not unique to artificial systems---after all, the human brain has various cognitive systems and modules with different (but interconnected) functions \citep{yeo2011organization}. We therefore think the best approach is to evaluate the system as a whole, including any built-in tools or modules. 

Measuring intelligence as a property of the system does have its drawbacks, though. For example, one difficult ramification is that the intelligence attributed to a given model depends on the harness built around it---analogous to concluding that a person becomes more intelligent when given access to a calculator or a computer. In addition, this approach raises questions about how to construct informative cognitive tests. When testing humans, access to external tools is typically restricted to maintain control over the testing conditions and target specific cognitive processes. If we allow AI systems to have access to any and all tools during evaluations, these tools could muddy the interpretation of the findings. For instance, imagine we want to test semantic memory for historical events. If the system can simply search the internet for information about these events, we are no longer measuring the system's memory---only its ability to search the internet. These are thorny issues that require consideration for each individual cognitive test. At the very least, any human baseline studies should ensure that participants are given access to the same tools we expect the AI systems to employ.

\subsection{Validation and iteration}
Like any science, the science of artificial general intelligence will be iterative in nature. The cognitive taxonomy described here is very much a starting point---it seems certain that AI systems will go on to develop cognitive capabilities that do not map neatly onto our taxonomy. Indeed, AI systems already possess some capabilities not found in humans, such as LiDAR perception and native image generation. Future iterations of this taxonomy will need to explore how to identify, characterize, and incorporate these emergent components of intelligence.

We also still have much to learn about how specific cognitive faculties are related to real-world performance. If a system lags behind humans in one or more of the 10 faculties, it would clearly demonstrate the system cannot match the generality of human intelligence. But what would a weakness in a given faculty mean when the system is deployed in the real world? There are good theoretical reasons to expect that each one of the 10 cognitive capacities identified here is important for different aspects of real-world performance. For example, a system that cannot plan would likely struggle with long-horizon, multi-step tasks, while a system deficient in social understanding is likely to perform poorly in situations that involve complex interactions with people. But empirical work is still needed to demonstrate these relationships and to further understand the importance of each cognitive capacity when it comes to practical, real-world tasks. 

\section{Conclusion}
The pursuit of Artificial General Intelligence represents a pivotal moment for humanity. By providing a clear, empirical framework rooted in the established science of human cognition, we aim to move the conversation around AGI from one of subjective claims and speculation toward a grounded, measurable scientific endeavor. Our Cognitive Taxonomy and evaluation protocol offer a way to map the jagged landscape of AI capabilities and track progress toward general intelligence.

\newpage
\section{Contributors and Acknowledgments}
\textbf{Contributors}
\begin{multicols}{4}
\raggedright
Alex Siegman\\
Alëna Aksënova\\
Anastasios Kementsietsidis\\
Arjun Narayanan\\
Ashwin Vaswani\\
Bernd Bohnet\\
Chrysovalantis Anastasiou\\
Claire Yao\\
Daniel McDuff\\
Dima Yeroshenko\\
Edward Loper\\
Ellie Pavlick\\
Evan Rosen\\
Georgi Karadzhov\\
Jed McGiffin\\
Joe Heyward\\
Julia Haas\\
Kartikeya Badola\\
Kate Lin\\
Laura Kampis\\
Martin Polacek\\
Martyna Płomecka\\
Mor Hazan Taege\\
Nicholas Cain\\
Nico Duduta\\
Rasmi Elasmar\\
Reut Aharony\\
Rivka Moroshko\\
Silvia Chiappa\\
Steve Zheng\\
Vik Sharma\\
Viorica Patraucean\\
Virginia Aglietti\\
William Cunningham\\
William Isaac\\
Xin Liu\\
Yao Yan\\
Yuan Yuan\\
\end{multicols}

This effort involved the contributions of many individuals across Google, including researchers, engineers, and operations staff. We gratefully acknowledge the dedication and hard work of each contributor on this effort. Contributors are listed in alphabetical order.

We are grateful for the invaluable feedback from Dharshan Kumaran, Joel Z Leibo, Ellie Pavlick, Mike Mozer, Martin Chadwick, Zoubin Ghahramani, Rohin Shah, Iason Gabriel, Lucy Cheke, and Jose Hernandez-Orallo.

\newpage
\section{Appendix: Cognitive taxonomy}
\label{sec:full-taxonomy}

\subsection{Perception}
\label{sec:perception}
The ability to take in and process sensory information from the world such as images, audio, and text \citep{harris2022sensation}. Perception allows a system to observe the environment and respond to its characteristics.\footnote{Input modalities are highly system dependent. A system may not need to perceive \textbf{all} modalities to behave intelligently, although each additional modality of information is likely to provide some benefits.} We organize this section by modality, and include aspects of language processing in the relevant modality.

\subsubsection{Visual perception}
The ability to take in and process visual information from light \citep{haber1973psychology, cornsweet1970visual}.

\paragraph{Low-level visual perception}
The ability to detect and identify surface-level features of visual information, such as light intensity, color, and contrast. \footnote{Here, we classify abilities as low-level perception when they involve “surface-level” processing with little semantic interpretation. Conversely, we classify abilities that require deeper, semantic processing as high-level visual perception. This is because there is evidence in humans that high-level and low-level perceptual processing can be dissociated (e.g. \citealt{farah1990visual}). However, the dividing line is somewhat fuzzy---perceptual abilities fall on a spectrum from those which require very little semantic processing (e.g., contrast detection) to those which are highly reliant on semantic processing (e.g., scene understanding).}

\begin{itemize}
    \item \textbf{Light detection:} The ability to detect the brightness of different parts of a visual scene \citep{gilchrist1999anchoring}.
    \item \textbf{Contrast (edge) detection:} The ability to detect contrast between light and dark areas of an image \citep{marr1980theory}. Critical for image segmentation.
    \item \textbf{Color detection:} The ability to distinguish between different colors in an image \citep{gegenfurtner2003cortical}.
    \item \textbf{Depth perception:} The ability to identify the distance to objects in a visual scene \citep{parker2007binocular}.
    \item \textbf{Motion detection:} The ability to detect movement in a dynamic visual scene \citep{borst1989principles}.
    \item \textbf{Shape detection:} The ability to identify and locate simple shapes in an image \citep{biederman1987recognition, todd2004visual}.
\end{itemize}

\paragraph{High-level visual perception}
The ability to process, interpret, and understand the semantic meaning of visual information \citep{ullman1996high}.
\begin{itemize}
    \item \textbf{Object recognition:} The ability to identify and categorize objects in a visual scene \citep{riesenhuber2000models}.
    \item \textbf{Visual spatial localization:} The ability to determine the spatial location of objects in a visual scene \citep{hollingworth2007object}.
    \item \textbf{Visual counting:} The ability to determine the number of objects or entities in a visual scene \citep{Trick1994}.
    \item \textbf{Static scene understanding (image understanding):} The ability to form a high-level understanding of events appearing in a static visual scene \citep{epstein2019scene}.
    \item \textbf{Dynamic scene understanding (video understanding):} The ability to form a high-level understanding of events occurring in a dynamic visual scene \citep{zacks2010brain}.
    \item \textbf{Word segmentation:} The ability to segment and identify letters and words in a visual scene \citep{norris2013models}.
    \item \textbf{Reading comprehension:} The ability to understand the meaning of language presented in a visual form \citep{kendeou2016reading}.\footnote{To demonstrate reading comprehension, a system needs to show mastery of at least one language. The ability to learn new languages is covered under learning, and the number of languages a system knows is a function of its semantic memory.}
\end{itemize}

\subsubsection{Auditory perception}
The ability to take in and process auditory information from sound waves \citep{warren1982auditory}.

\paragraph{Low-level Auditory perception}
The ability to extract surface-level features of auditory information such as loudness, pitch, and spatial location.
\begin{itemize}
    \item \textbf{Loudness detection:} The ability to identify the loudness (volume) of a sound \citep{fechner1860elemente}.
    \item \textbf{Pitch detection:} The ability to identify the pitch of a sound \citep{oxenham2012pitch}.
    \item \textbf{Sound discrimination:} The ability to differentiate components of an auditory scene occurring simultaneously, such as different sounds or speakers \citep{bizley2013perception}.
    \item \textbf{Sound localization:} The ability to identify the direction of a sound and the distance to its source \citep{blauert1996spatial}.
    \item \textbf{Speech segmentation:} The ability to separate a stream of sound information into separate words \citep{brent1999speech}.
    \item \textbf{Rhythm perception:} The ability to identify patterns or rhythms in sound information \citep{grahn2012neural}.
\end{itemize}

\paragraph{High-level Auditory perception}
The ability to process, interpret, and understand auditory information.
\begin{itemize}
    \item \textbf{Speech comprehension:} The ability to understand the meaning of spoken language \citep{diehl2004speech}.
    \item \textbf{Speaker identification:} The ability to recognize and distinguish individual speakers \citep{mcdermott2009cocktail}.
    \item \textbf{Sound recognition:} The ability to recognize different types of sounds.
    \item \textbf{Auditory scene understanding:} The ability to understand and interpret events occurring in an auditory scene \citep{bregman1994auditory}.
    \item \textbf{Listening comprehension:} The ability to understand the meaning of language presented in auditory form \citep{friederici2002towards}.\footnote{As with reading comprehension, to demonstrate listening comprehension a system needs to show mastery of at least one language. The ability to learn new languages is covered under learning, and the number of languages a system knows is a function of its world knowledge.}
\end{itemize}

\subsubsection{Text perception}
The ability to process, interpret, and understand information presented as text.

Unlike humans, who perceive text only through other modalities (e.g. through vision via reading, or through touch via braille reading), today's AI systems perceive text as an entirely separate modality via tokenization and embedding. These processes are flawed and lossy much like other forms of perceptual processing \citep{shin2024large}---as illustrated by systems' difficulties with spelling and their brittleness to perturbations in text inputs.

Although text perception is not strictly a fundamental capability that humans have, textual information is central to today's society and is an essential part of how today's systems are trained to understand language and the world. The ability to bypass vision and directly perceive text-based language is a fascinating and unique property of today's AI systems. Language models have demonstrated that this ability can enable remarkable performance on a range of tasks, so we think it is worthy of consideration and measurement. Of course, the lack of a direct equivalent in humans raises questions about how text perception should factor into judgments about AGI. In one sense, this ability can be thought of as an "easier" version of reading, without the need to segment the visual information into letters and words. Therefore, we think that an AGI should, at a minimum, be able to comprehend a given text at least as well as a human could if presented with that text in visual form.

\paragraph{Low-level text perception}
The ability to extract surface-level features from text inputs such as letters and words.

\begin{itemize}
    \item \textbf{Symbol discrimination:} The ability to discriminate between different symbols in a stream of characters (e.g. letters and numbers).
    \item \textbf{Word segmentation:} The ability to segment a stream of characters into words.
\end{itemize}

\paragraph{High-level text perception}
The ability to process, interpret, and understand text information.

\begin{itemize}
    \item \textbf{Language comprehension:} The ability to understand the meaning of language presented as text \citep{traxler2006handbook}.
    \item \textbf{Code comprehension:} The ability to understand the meaning and function of written code. 
\end{itemize}

\subsubsection{Other modalities of perception}
There are many other types of perception seen in humans or other animals that may provide utility for AI systems. It seems likely that each additional modality of perception will provide some useful information to systems that can help them navigate the real world. For example, touch \citep{de2020somatosensation}, smell \citep{stevenson2009initial}, and temperature perception \citep{jung2023temperature} could all prove highly useful for a system designed to help with cooking or manufacturing. Still, the relative importance of these modalities for intelligent behavior remains unclear, so in this initial version of the taxonomy we do not include them in detail. These modalities are nonetheless worth considering, both for researchers building new systems and those building model evaluations.

\subsubsection{Multi-sensory integration}
The ability to integrate information from multiple modalities together. Because different perceptual modalities provide complementary information about features such of the environment, such as form, location, size, and speed of movement, multi-sensory integration helps enable joint processing, reasoning, and planning across the information \citep{zmigrod2013feature}. This ability should be measured for each combination of modalities.

\subsection{Generation}
The ability to generate outputs such as text, speech, or actions such as motor movements, computer control actions, or tool use calls. Output generation is critical for a system's ability to communicate and complete tasks effectively.

The ability to generate high-quality outputs can be at least partly decoupled from the ability to decide which output to generate. In other words, output generation captures a system's \textit{execution} ability, not its ability to reason and plan about which actions to attempt.

\subsubsection{Text generation}
The ability to generate text outputs. In humans, text generation is indirect, typically mediated through motor actions (via writing or typing), but in AI systems text can be a direct output.

\paragraph{Natural language generation}
The ability to generate natural language in text form. Typically known as language \textit{production} in cognitive science \citep{Garrett1988, Pickering_Garrod_2013}.
\begin{itemize}
    \item \textbf{Grammatical correctness:} The ability to produce grammatically correct sentences.
    \item \textbf{Lexical selection:} The ability to select appropriate words to convey rich meaning.
\end{itemize}

\paragraph{Code generation}
The ability to generate structurally and syntactically correct code (see \citealt{FEDORENKO2019525} for a discussion).

\subsubsection{Audio generation}
The ability to produce audio outputs (i.e., sound).

\paragraph{Speech generation}
The ability to generate natural and expressive speech \citep{traxler2006handbook}. 

\begin{itemize}
    \item \textbf{Clarity:} The ability to clearly pronounce words and phonemes.
    \item \textbf{Grammatical correctness:} The ability to produce grammatically correct sentences.
    \item \textbf{Lexical selection:} The ability to select appropriate words to convey rich meaning.
    \item \textbf{Prosody control:} The ability to control the rhythm of speech such as pitch, stress, speed and intonation.
    \item \textbf{Emotional expression:} The ability to express a wide range of emotions through variation in prosody.
\end{itemize}

\subsubsection{Action generation}
The ability to generate actions that manipulate the environment.

\paragraph{Motor control}
The ability to control a body or robotic actuators \citep{Nishikawa2007Neuromechanics}.
\paragraph{Computer control}
The ability to produce computer control actions such as key presses and mouse movements \citep{smith1999aging}.
\paragraph{Tool use}
The ability to use external objects, systems, or resources to assist in the pursuit of goals \citep{baber2003cognition}.

\subsubsection{Thought generation}
The ability to generate internal thoughts which can be used to guide decisions \citep{holyoak1993thinking}. Thoughts could take the form of language, images (akin to human visual imagination), or could take a more abstract form.

By definition thoughts are internal in nature and may be difficult or impossible to evaluate. However, conscious thought is critical for human problem solving and there is substantial evidence for its value in AI systems \citep{openai2024learningtoreason, comanici2025gemini}, so evaluating the features and quality of a system's thoughts will almost certainly be enlightening.

\subsection{Attention}
The ability to focus cognitive resources on specific aspects of perceptual stimuli, information, or thoughts \citep{treisman1969strategies, styles2006psychology}. This is critical when a system is faced with a complex environment and has limited cognitive resources.

There is a delicate balance between the need to avoid distraction by narrowly focusing on information that is important for current goals and the need to stay attentive to the wider environment to respond when unexpected changes or stimuli appear \citep{burgoyne2020attention, engle2018working}.

\subsubsection{Attention capacity}
The amount of information a system can focus on simultaneously. This capacity may be different for different modalities or types of information (e.g., text, images, audio; \citealt{cowan2005capacity, fritz2007auditory}).

\subsubsection{Selective attention / Attentional control}
The ability to selectively focus on information that is relevant to current goals and ignore information that is goal-irrelevant \citep{theeuwes1991exogenous}. This active, top-down form of attention is important for effective goal-driven behavior and is often considered an executive function.

\paragraph{Sustained attention}
The ability to maintain focus on goal-relevant information over time \citep{sarter2001cognitive, esterman2019models}.
\paragraph{Perceptual inhibition}
The ability to ignore distracting or goal-irrelevant perceptual information \citep{van2020inhibition}.
\paragraph{Attention shifting}
The ability to actively shift attention from one location or piece of information to another \citep{brown2016attentional}.

\subsubsection{Stimulus-driven attention}
The ability for attention to be directed in a "bottom-up" way toward new stimuli or environmental changes \citep{katsuki2014bottom, awh2012top}. This is crucial for quickly identifying situational shifts that require a response. 

\subsection{Learning}
\label{sec:learning}
The ability to acquire new knowledge, skills, or behaviors through experience, study, or instruction. Learning is vital for a system’s ability to adapt to new situations or environmental changes \citep{morand2017learn, ginsburg2010evolution}.\footnote{For many current systems, learning occurs only during training or in-context. However, for truly robust and adaptive behavior, AI systems should be able to learn (and retain) new knowledge and skills over time (e.g., as part of a continuous learning process).} Here, we enumerate several important kinds of learning an AGI should be able to employ.

\subsubsection{Concept formation}
The ability to abstract the key features of objects, events, and ideas to form categories, concepts, schemas, and scripts \citep{gershman2010learning, Bruner1986Study, goodman2008rational, tenenbaum2011grow}. Important for generalization.

\subsubsection{Associative learning}
The ability to learn the relationships between events, objects or stimuli that appear together \citep{shanks1995psychology}.

\subsubsection{Reinforcement learning (Operant conditioning)}
The ability to learn based on the consequences (rewards and punishments) of specific actions or situations \citep{sutton1998reinforcement, skinner1963operant}.

\subsubsection{Observational learning}
The ability to learn by watching others perform a skill or task \citep{bandura2008observational}.

\subsubsection{Procedural learning}
The ability to learn skills, action patterns, or tasks through performance or practice \citep{cohen1980preserved}.\footnote{In humans, procedural learning can be dissociated from the ability to learn facts and other explicit information \citep{willingham1989development}.}

\subsubsection{Language learning}
The ability to learn new language-related information, such as syntax and vocabulary \citep{chomsky1965aspects, pinker1994language, saffran2003statistical, bates1976language}. Includes natural languages as well as coding languages and tool use frameworks.

\subsection{Memory}
The ability to keep track of information over time \citep{squire1999memory}. Memory and learning are closely linked---the distinction is that learning is focused on the acquisition of new knowledge, whereas memory is concerned with the ability to maintain that knowledge over time. Evaluating memory typically involves testing a system's pre-existing knowledge as well as its ability to store and retrieve newly-learned information. The quality of a system's memory can be considered in terms of how much information a system can remember, how long the information can be maintained, and/or how accurate the memories are.\footnote{Most research on human memory is focused on understanding the specific memory mechanisms of the brain, such as “long-term memory” and “short-term memory”. But these mechanisms are specific to humans and may not be relevant to AI systems. For this reason, we think it is important to remain agnostic to the underlying memory implementation and focus on evaluating \textit{how well} a system can keep track of information over time.}

Given how closely linked learning and memory are, it may be difficult to separate them when evaluating system capabilities. However, there are at least some failure modes that are specific to one or the other. For example, a failure to update semantic knowledge despite being able to successfully recall already stored knowledge would be considered a failure of learning, while forgetting information over time that was initially successfully learned would be a failure of memory.

\subsubsection{Semantic memory (World knowledge)}
\label{sec:semantic-memory}
The ability to keep track of facts and other general information not tied to a specific episode \citep{yee2013semantic, kumar2021semantic}.

\paragraph{Commonsense knowledge}
Knowledge about the fundamental rules, properties, and characteristics of the world. Includes:
\begin{itemize}
    \item General knowledge
    \item Causal knowledge
    \item Temporal knowledge
    \item Spatial knowledge
    \item Intuitive physics knowledge
\end{itemize} 

\paragraph{Domain knowledge}
Specialist knowledge about specific subjects or domains \citep{ericsson2018cambridge}. For example:
\begin{itemize}
    \item Linguistic knowledge
    \item STEM knowledge
    \item Coding knowledge
    \item Legal knowledge
    \item Financial knowledge
    \item Medical knowledge
    \item Historical knowledge
    \item Social-cultural knowledge
\end{itemize} 

\subsubsection{Episodic memory}
The ability to keep track of information relating to specific events, including the sensory information (images, sounds, etc.) associated with those events \citep{tulving1972episodic, conway2009episodic}.\footnote{The ability to remember sensory information is sometimes separated from the ability to remember factual details about specific events. The relationship between these aspects of event memory is complex, so for practical purposes we use episodic memory to refer to both together.}

\paragraph{Sensory memory}
The ability to keep track of sensory information from specific events (e.g. visual, auditory, or olfactory information).

\paragraph{Temporal memory}
The ability to keep track of the sequence and temporal relationships between events.

\paragraph{Spatial memory}
The ability to keep track of the spatial relationships between objects from specific events (e.g. their positions, orientations, and movements).

\subsubsection{Procedural memory}
The ability to keep track of the action and output patterns needed to perform skills \citep{cohen1980preserved}.

\subsubsection{Prospective memory} 
The ability to remember to perform a planned action when a specific cue arises, such as a moment in time or in response to a specific event (e.g. “When I go to the store, I should remember to get milk”; \citealt{mcdaniel2007prospective}). 

\subsubsection{Forgetting}
The ability to remove outdated, wrong, or irrelevant information from memory. This could involve information compression or wholesale pruning of specific memories. Forgetting is important for efficient storage and retrieval \citep{bjork1989retrieval}.

\subsection{Reasoning}
The ability to draw valid conclusions by applying logical principles or drawing inferences \citep{leighton2003nature, rips1990reasoning}. Typically, automatic pattern matching would not be considered reasoning.

\subsubsection{Deductive (logical) reasoning}
The ability to reason from a set of premises, rules or facts to reach a logical conclusion \citep{johnson1999deductive}. A defining feature of deduction is a lack of ambiguity---as long as one can be sure the premises are true, then correctly applied deduction will lead to conclusions that are certain. Deduction requires a grasp of logical concepts such as negation, “AND”, “OR”, “XOR”, and “ALL”.

\subsubsection{Inductive reasoning}
The ability to draw general conclusions based on a specific set of facts, information, or observations (e.g., “The sun has risen every day so far; therefore, the sun will rise tomorrow”).

Unlike deductive reasoning, these general conclusions are probabilistic, not certain. As a result, induction can occasionally lead us astray. However, in a world in which there is constant uncertainty, inductive reasoning is essential to avoid constant paralysis \citep{heit2000properties}.

\subsubsection{Abductive reasoning}
The ability to make inferences about the best or most likely explanation for a set of observations. Much like inductive reasoning, abductive reasoning is uncertain. A key feature of abduction is that it involves the generation of new explanatory hypotheses, rather than just generalizing from a set of observations \citep{bhagavatula2020abductive}.

\subsubsection{Analogical reasoning}
The ability to identify similarities between situations or concepts and to use those similarities to draw conclusions about unknown properties of one based on known information about the other \citep{gentner2018analogical,alexander2016relational}.

\subsubsection{Mathematical reasoning}
The ability to perform mathematical calculations and operations \citep{gilmore2018introduction}. Includes basic arithmetic and algebra.

\subsection{Metacognition}
The knowledge a system has about its own cognitive processes, and its ability to monitor and control those processes \citep{dunlosky2009metacognition, tarricone2011taxonomy}.

\subsubsection{Metacognitive knowledge}
Metacognitive knowledge is a system’s self-knowledge about its own abilities, limitations, knowledge, learning processes, and behavioral tendencies \citep{flavell1979metacognition, tarricone2011taxonomy}. In some ways, metacognitive knowledge could simply be considered a special case of world knowledge.

\paragraph{Knowledge of limitations}
Knowledge of one's own capabilities and limitations \citep{fleming2014measure, fleming2017self}. 
\paragraph{Knowledge of learning processes}
Knowledge of one's own learning processes and the factors that can assist or impair learning \citep{wang2021meta, binz2024meta}.
\paragraph{Metamemory (Knowledge of knowledge)}
Knowledge about the information stored in memory and about the processes involved in storing or retrieving that information \citep{nelson1990metamemory}.
\paragraph{Knowledge of behavioral patterns}
Knowledge of one's own tendencies and behavioral patterns \citep{grant2001rethinking, vazire2010self}.

\subsubsection{Metacognitive monitoring}
The ability to monitor the state of cognitive processes (e.g., evaluating the state of learning or current performance) \citep{dunlosky2009metacognition, nelson1990metamemory}.

\paragraph{Confidence calibration}
The ability to accurately estimate the likelihood that an action will be successful or that a response will be correct \citep{harvey1997confidence, fleming2014measure, yeung2012metacognition}
\paragraph{Judgments of learning}
The ability to monitor progress when learning new information \citep{arbuckle1969discrimination, rhodes2016judgments}
\paragraph{Error monitoring}
The ability to notice when errors are made \citep{yeung2012metacognition}.
\paragraph{Source judgments}
The ability to judge where a piece of information was generated or learned from \citep{johnson1993source, mitchell2000source}

\subsubsection{Metacognitive control}
A model's ability to utilize insights from metacognitive knowledge and monitoring to adjust cognitive processes or strategies \citep{nelson1990metamemory, son2002relation, botvinick2007conflict}. Sometimes considered an executive function.

\paragraph{Error correction}
The ability to adjust action patterns or strategies to correct errors \citep{metcalfe2017learning}.
\paragraph{Learning strategy selection}
The ability to select appropriate learning strategies based on metamemory and judgments of learning (e.g. terminating study of well-learned information to focus on information that has not yet been well-learned) \citep{dunlosky2013improving}.

\subsection{Executive functions}
Higher-order cognitive abilities that enable goal-directed behavior by regulating and orchestrating thoughts and actions \citep{diamond2013executive}.

\subsubsection{Goal setting and maintenance}
The ability to set and maintain goals to organize and direct action \citep{dickinson1994motivational, buschman2014goal}. 

\subsubsection{Planning} 
The ability to formulate sequences of future actions to achieve specific goals \citep{owen1997cognitive}. Planning is a key part of solving multi-step or long-term problems. The process of planning can be broadly construed as building up (and pruning) some kind of decision-tree \citep{mattar2022planning}.

\subsubsection{Inhibitory control}
The ability to change, withhold, or suppress learned or habitual responses in favor of more controlled, goal-appropriate ones \citep{miyake2000unity, bari2013inhibition}. 

\subsubsection{Cognitive flexibility}
\label{sec:cognitive-flexibility}
The ability to switch between different tasks, concepts, or ways of thinking \citep{braem2018getting}.

\subsubsection{Conflict resolution}
The ability to manage and resolve conflicting information, contradictory goals, or competing responses to select an appropriate action \citep{botvinick2001conflict, veen2006conflict}. Not to be confused with social conflict resolution.

\subsubsection{Working memory}
The ability to manipulate information internally in service of a goal (e.g. performing intermediate calculations while solving a problem or mentally rotating an image to consider it from a different perspective) \citep{baddeley1992working}. Although the name would suggest this ability is a subset of memory, in truth working memory involves the coordination of multiple faculties including memory, attention, and sometimes reasoning \citep{engle2002working}.

\subsection{Problem solving}
\label{sec:problem-solving}
As the name suggests, this broad ability refers to the ability to solve problems and overcome obstacles \citep{MayerWittrock2006}. This is a composite ability that relies heavily on planning, reasoning, and in-context learning. Problem solving requires:
\begin{itemize}
    \item Understanding and representing the problem (e.g. via perception and abstraction)
    \item Identifying and retrieving relevant knowledge (e.g., facts, analogous episodes, meta-knowledge about effective problem-solving strategies).
    \item Breaking down the problem into sub-goals
    \item Planning a sequence of actions to take 
    \item Executing the plan (e.g. via reasoning + output generation)
\end{itemize}

It would be impossible to enumerate all types of problems that humans are able to solve, but we describe several important types below.

\subsubsection{Fluid reasoning}
The ability to identify patterns and apply them to solve novel problems. Relies on a mix of deductive, inductive, and abductive reasoning \citep{cattell1943measurement, kent2017fluid}.

\subsubsection{Mathematical problem solving}
Applying mathematical concepts and techniques to solve problems \citep{schoenfeld1985mathematical}.

\subsubsection{Algorithmic problem solving}
Solving logical problems using algorithmic techniques---a key part of writing code.

\subsubsection{Commonsense problem solving}
Solving real-world problems by applying general knowledge and everyday understanding of the world \citep{BrachmanLevesque2022}.

\paragraph{Temporal problem solving}
Understanding and reasoning about temporal concepts such as the order and duration of events, time-based relationships, and temporal constraints.

\paragraph{Spatial problem solving}
Reasoning about the relationships between objects in space, including their positions, orientations, and movements.

\paragraph{Causal problem solving}
Reasoning about cause-and-effect relationships between events or entities.

\paragraph{Intuitive physics}
Reasoning about basic physical principles and how objects behave in the world. Includes concepts such as:
\begin{itemize}
    \item Object permanence
    \item Gravity
    \item Momentum
    \item Force
\end{itemize}

\subsubsection{Knowledge discovery} 
The ability to generate novel hypotheses, experiments, and solutions to scientific questions \citep{klahr2000exploring, dunbar2001scientific, nersessian2002cognitive}. 

\subsection{Social cognition}
The ability to process and interpret social information and to respond appropriately in social situations. Crucial for interacting with people or other systems \citep{higgins1987social}.

\subsubsection{Social perception}
The ability to interpret social cues based on perceptual information such as facial expressions, tone of voice, or body language \citep{Tajfel1962}.

\subsubsection{Theory of mind}
The ability to reason about the mental states of others, including beliefs, desires, emotions, intentions, expectations, and perspectives. Theory of mind is important for the ability to predict and explain others’ behavior \citep{frith2005theory, leslie2004core}.

\subsubsection{Social skills}
The ability to recognize, understand, and act according to social norms or expectations \citep{chung2016social}.

\paragraph{Cooperation}
The ability to work together with others toward common goals \citep{rand2013human}.

\paragraph{Negotiation}
The ability to work together with others toward goals that are misaligned or in conflict \citep{bazerman2000negotiation}.

\paragraph{Deception}
The ability to mislead others by hiding or disguising intentions or actions in order to achieve goals \citep{spence2004cognitive}. Could be considered a harmful capability depending on the context.

\paragraph{Persuasion}
The ability to influence others’ attitudes, beliefs, or behaviors \citep{wood2000attitude}. Could be considered a harmful capability depending on the context.

\bibliography{main}

\end{document}